\pdfoutput=1

\documentclass[11pt]{article}

\usepackage{EACL2023}

\usepackage{times}
\usepackage{latexsym}

\usepackage[T1]{fontenc}

\usepackage[utf8]{inputenc}

\usepackage{microtype}

\usepackage{inconsolata}
\usepackage{graphicx} 

\usepackage{subcaption}
\usepackage{xspace}

\newcommand{\dataset}{\textbf{\texttt{LEGIT}}\xspace}
\newcommand{\trocr}{\textsc{TrOCR}\xspace}
\newcommand{\clip}{\textsc{CLIP}\xspace}
\newcommand{\beit}{\textsc{BEiT}\xspace}
\newcommand{\detr}{\textsc{DETR}\xspace}
\newcommand{\imgdot}{\textsc{ImgDot}\xspace}

\usepackage{amsmath}
\usepackage{multirow}
\usepackage{booktabs}

\title{Learning the Legibility of Visual Text Perturbations}

\author{
Dev Seth$^\dagger$\quad Rickard Stureborg$^\dagger$ \quad Danish Pruthi\thanks{~~~Work done while at Carnegie Mellon University.} \quad Bhuwan Dhingra$^\dagger$ \\
$^\dagger$ Duke University \\
{\texttt{\{ds447, rs541, bd149\}@duke.edu}}\\
{\texttt{danish@hey.com}}
}

\begin{document}
\maketitle

\setlength{\abovedisplayskip}{3pt}
\setlength{\belowdisplayskip}{3pt}

\begin{abstract}
   Many adversarial attacks in NLP perturb 
   inputs to produce visually similar strings 
   (‘ergo' $\rightarrow$ ‘$\varepsilon$rgo') 
   which are legible to humans but degrade model performance. 
   Although preserving legibility is a necessary condition for text perturbation, 
   little work has been done to systematically characterize it; 
   instead, legibility is typically loosely enforced via 
   intuitions around the nature and extent of perturbations. 
   Particularly, it is unclear to what extent can inputs be 
   perturbed while preserving legibility, 
   or how to quantify the legibility of a perturbed string. 
   In this work, we 
   address
   this gap by 
   learning models that predict the legibility of a perturbed string, 
   and
   rank candidate perturbations based on their legibility. 
   To do so, we collect and release \dataset, 
   a human-annotated dataset comprising the
   legibility of visually perturbed text. 
   Using this dataset, we build both text- and vision-based 
   models which achieve up to $0.91$ F1 score in predicting whether an input is legible, 
   and an accuracy of $0.86$ in predicting which of two given perturbations is more legible. 
   Additionally,
   we discover that 
   legible perturbations from the \dataset dataset 
   are more effective at lowering the performance of NLP models
   than best-known attack strategies, 
   suggesting that current models may be vulnerable to a broad 
   range of perturbations 
   beyond what is captured by
   existing visual attacks.\footnote{Data, code, and models are available at \url{https://github.com/dvsth/learning-legibility-2023}.}
\end{abstract}

\section{Introduction}
\begin{figure}[t]
    \centering
    \includegraphics[width=2.9in]{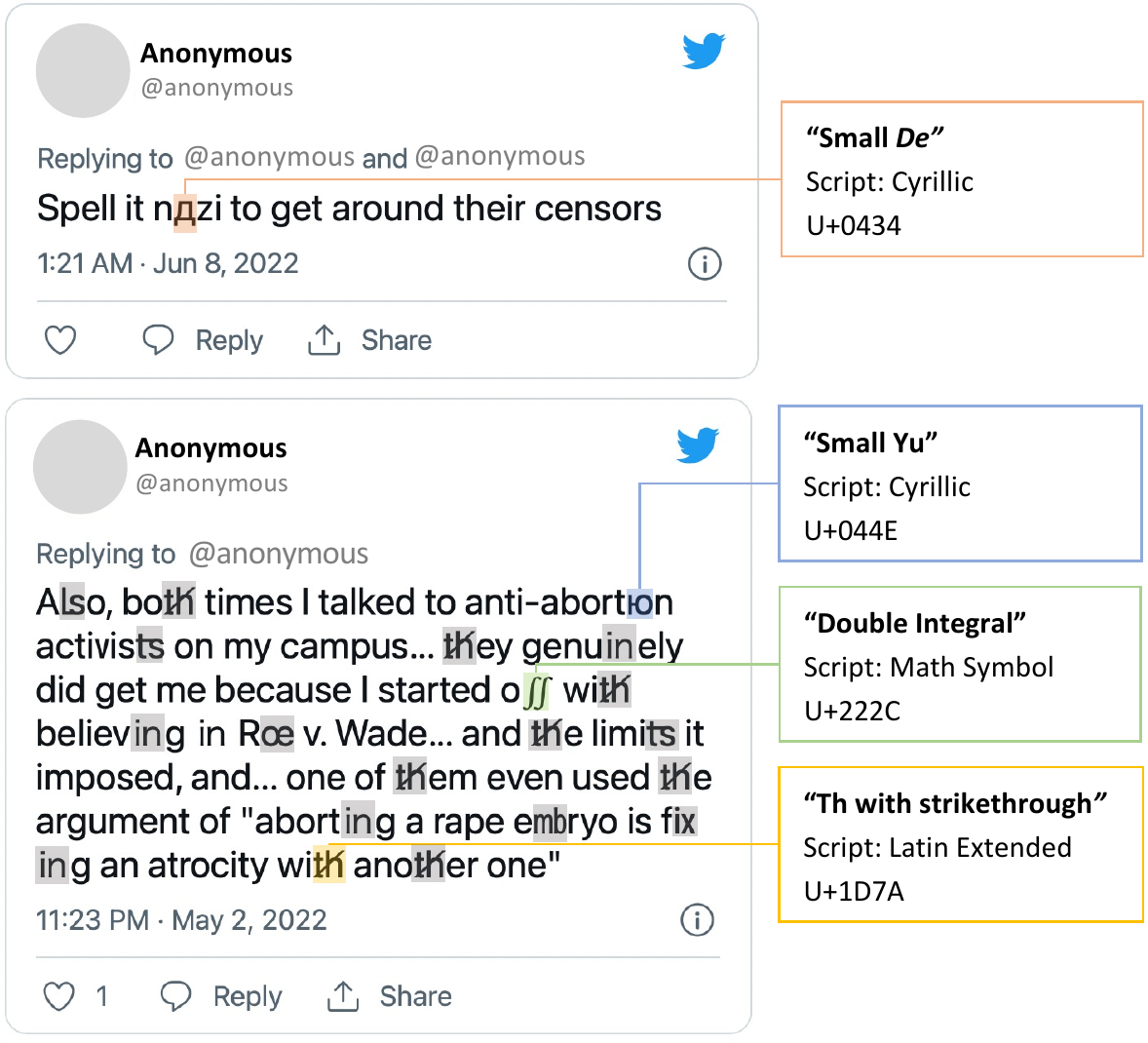}
    \caption{Visual attacks in the wild. Examples of Twitter users manipulating their tweets to evade the platform's `sensitive content' detection algorithms.
    }
    \label{fig:manipulated_examples}
\end{figure}

To manage the increasing demand 
for content moderation---e.g., detecting spam or toxic/hateful
content on online platforms---organizations have
turned to 
machine learning solutions. 
In response, users often 
resort 
to 
manipulating text to evade 
detection, removal, or search.
For instance, 
hateful 
comments 
often 
comprise of visually similar characters 
to avoid automatic filtering \cite{le_perturbations_2022}. 
Since people read text visually, 
the manipulated content can still be easily understood and 
harm its target audience.
These attacks started with
simple ASCII substitutions like he\underline{11}o 
(colloquially referred to as ``leetspeak"),
but have evolved into complex manipulations %
utilizing characters from different Unicode scripts \cite{flamand2008deciphering, raymond1996new}.
\autoref{fig:manipulated_examples} shows two such examples.

Unlike computer vision 
where there is an established 
notion of 
what constitutes an imperceptible perturbation
(typically defined via the $\ell_\infty$ distance), 
most perturbations in text are perceptible. 
However, as long as the perceptible 
manipulations remain \textit{legible}, 
the message could have its intended effect.
The legibility of a text is determined by whether or not a literate
person can decipher the altered words.
The degree to which a piece of text can be perturbed, while maintaining
legibility, depends on a multitude of factors such
as its context, similarity to the original content, the positions of the perturbations, 
the background knowledge of the reader, etc. However,
many adversarial attacks 
enforce legibility only loosely based on 
intuitions about the nature of the attacks, e.g., that changing $1$-$2$ characters in a sentence does not impact its legibility~\cite{belinkov2017synthetic, pruthi_combating_2019}.

In this work, we instead propose to learn 
the legibility of visual perturbations, 
by developing text- and vision-based models
trained on legibility annotations from human subjects. 
The current focus of research on adversarial attacks 
is to find minimal perturbations required to break NLP models,
and several recent findings suggest that models remain brittle to such perturbations~\cite{eger_text_2020, dionysiou_unicode_2021, pruthi_combating_2019}. 
In contrast, our work attempts to uncover the space of \textit{all legible
perturbations} that we need to defend against. 
Towards our goal of characterizing the limits of legibility of perturbed texts,
we make the following contributions:

First, 
we crowdsource 
human judgments about the legibility of different perturbations: specifically, we show annotators two perturbed versions of the same word and ask them which one, if any, they find more legible. 
Our perturbation strategy considers 
substituting letters in the word 
with Unicode characters drawn from a large subset of the Basic Multilingual Plane covering
over $100$ scripts from around the world.\footnote{We consider $12,287$ Unicode characters from codepoints 0x0000 to 0x2fff.
}
In total, we collect $30,320$ annotations, one each for $14,643$ and $3,332$ instances
in the training and validation sets, respectively, and three each for the $4,113$ instances in the test set.
Using these preferences, we define a \emph{pairwise legibility ranking task} as well as
a \emph{binary legibility classification task}.
While the former allows making inferences about which
candidate perturbation
is \textit{most} legible,
the latter allows filtering out illegible perturbations
altogether. For each task, we identify a \emph{hard} subset of the collected data, which includes fine-grained comparisons expected to be more challenging for annotators and models alike.

Second, we use the labeled data to train models which predict
the degree of legibility of a perturbed text. Specifically, we fine-tune pretrained vision
\cite[TrOCR;][]{li_trocr_2021}
and text-based
\cite[ByT5;][]{xue_byt5_2021}
models on the ranking and classification tasks.
We find that TrOCR trained in a multi-task setup on both
tasks achieves the best performance with $0.91$
F1 score on the classification task and $0.86$ accuracy on the ranking task.
Interestingly, we find that the purely text-based
ByT5 also achieves competitive performance on the classification
task with $0.89$ F1, suggesting that its pretrained byte representations
encode aspects of visual similarity between Unicode characters.
Further, we find that models have high F1 scores on the subset
of data with high inter-annotator agreement: TrOCR
achieves a $0.96$ F1 score on test cases where all three
annotators agree. 
We also note that
legibility is a complex phenomenon---it doesn't 
correlate trivially with the distance of the perturbation from the 
original text or the number of letters substituted.\footnote{A logistic regression model using 
these as features only agrees $56.7\%$ of the time with authors' legibility assessment.}

Third, we consider a word-level \emph{perturbation recovery} task,
which involves inferring the original word from its perturbed version.
We evaluate
GPT-3 \cite{GPT3} on this task, comparing its performance on legible perturbations from our perturbation strategy versus those generated by VIPER, a \emph{VIsual PERturber} method proposed by \citet{eger_text_2020}.
We find that GPT-3 has a lower accuracy
in recovering perturbations from our perturbation strategy,
despite VIPER providing no guarantees on legibility.
Additionally, we apply our findings to the important task of toxicity classification.
We perturb a subset of the dataset using our perturbation strategy and find that it degrades the SOTA Detoxify \cite{Detoxify} classifier more than existing VIPER attacks.
These findings demonstrate that existing attacks do not comprehensively cover the space of legible perturbations that can degrade model performance.

\section{Related Work}
\paragraph{Adversarial Attacks for NLP.}
A challenge in defining adversarial examples for text
lies in characterizing the space of \textit{equivalent}
inputs to a training or test example which preserves the target label.
While early work focused on adding distracting text to fool question
answering systems \cite{jia2017adversarial},
recent work utilizes more general strategies applicable to many
tasks \cite{li_textbugger_2019,morris_textattack_2020,jin2020bert}.
Many of these can be categorized as \textit{word}-level synonym substitutions
\cite{alzantot_generating_2018,garg_bae_2020,li2020bert},
or \textit{character}-level legibility-preserving substitutions~\cite{ebrahimi_adversarial_2018, pruthi_combating_2019}.
Most attacks in either category are perceptible in that readers
of the text can identify that it has been transformed,
except for one notable exception where invisible characters and 
near-identical characters
are used to render strings indistinguishable
from the original~\cite{boucher_bad_2021}. 
Attacks based on visual similarity of characters have also been previously
considered by
\citet{eger_text_2020} who propose three attack strategies: ICES (based on rendered glyph similarity), DCES (based on bag-of-words textual similarity of Unicode codepoint descriptions), and ECES (based on adding diacritics to base characters). For ICES, they compute similarity by comparing raw pixel values of the renderings,
which we improve upon here by utilizing a pretrained Optical Character
Recognition (OCR) model.
This produces a `smarter' set of visual neighbors: e.g., mirror images of letters, scaled versions of letters (like O vs $^\circ$) etc., which go beyond simple accents or modifiers. We also report in-depth comparisons between our perturbation strategy and the ECES and DCES approaches in \autoref{sec:results}.

\paragraph{Legibility of Perturbed Inputs.}
Among character-level perturbation attacks,
legibility 
has only been loosely enforced based on intuitions about the nature 
and the degree of manipulations. This often results in conservative substitutions which only represent a lower bound on the space of all legible perturbations. 
For instance, \citealp{pruthi_combating_2019} limit the attack to
only $1$-$2$ character changes (e.g., substitutions, deletions or additions)
per input example; similarly, \citealp{ebrahimi_adversarial_2018} propose an attack strategy which specifically minimizes the number of character manipulations required in order to render the output legible.
Attacks based on visual similarity 
usually constrain their attack surface to
inputs which are above a threshold similarity (in pixel or embedding space)  
to the original input~\cite{eger_text_2020, eger_hero_2020, dionysiou_unicode_2021}.
In this work, by contrast, we directly 
address the question of what constitutes legible perturbations, with the aim of learning a grounded definition of legibility rather than assuming one \emph{a priori}.

\section{Legibility Tests}

We adopt a supervised learning approach for determining
the legibility of perturbed texts.
In this section, we describe the process used
for collecting the \dataset dataset (which stands for \textbf{LEGI}bility \textbf{T}ests) and in the next
section we describe the modeling techniques used for
predicting the legibility score and ranking different candidate perturbations.

Our setting involves one-to-one character substitutions at the word level, i.e., given a word (and no other context), we consider perturbations where each letter in the word may be replaced by a Unicode codepoint in 0x0000-0x2fff. Moreover, the substitutions are mutually independent and do not depend on the context of the other letters.

\subsection{Perturbation Process}
\label{sec:perturbation}
To generate perturbations for the data labeling task, we replace a subset of characters in a word with visually similar counterparts.
Specifically, given a word $w$,
we first randomly select a fraction $n \in [0, 1]$ of characters in that
word to corrupt.
Then, each of the chosen characters is replaced by its nearest neighbor
at rank $k$
in the embedding space generated by a model $\mathcal{M}$
which encodes characters into visual features.
Hence, there are three parameters involved in the perturbation
process $\phi = \{n, k, \mathcal{M}\}$.

We experiment with several models to encode characters into visual
features, all based on renderings of the Unicode codepoints into
images.
To keep visual representations consistent across models, we use GNU Unifont, rendering each glyph separately in $144$px font size with black color, on a $224\times224$px white background.\footnote{
Glyphs were rendered by the Pillow library~\cite{clark2015pillow}.
}
Given the rendering, we compare $5$ models to
encode the features.
Three are transformer-based: \trocr (`base') \cite{li_trocr_2021}, \clip (`vit-base-patch32') \cite{radford_learning_2021}, and \beit (`base-patch16-224-pt22k-ft22k').
One employs convolutional as well as transformer networks: \detr \cite{carion_end--end_2020}.
The fifth model is a simple baseline: \imgdot, which uses the (flattened) bitmap of a rendered character as its embedding vector.
In preliminary experiments, $400$ perturbed pairs were generated, with each pair using the same settings for $k, n$ but using different models. The authors then independently ranked perturbations each pair based on their legibility. \detr- and \beit-generated perturbations were ranked above other models' perturbations $23\%$ and $41\%$ of the time, respectively, whereas \clip and \imgdot perturbations were preferred over others in $66\%$ and $73\%$ of cases. Hence, \detr and \beit were excluded from further experiments. \trocr was included later, after verifying that it was preferred $\approx 50\%$ of the time against both \clip and \imgdot.

For each of the chosen models, we compute the pairwise cosine distances between the model's embedding vectors for all Unicode codepoints in the range 0x0000--0x2fff (excluding invalid or empty codepoints), and use these distances to find the nearest-neighbors for each character.
Then, to perturb a given word $w$ using the parameters $\phi=\{k,n,\mathcal{M}\}$, we first pick 
$\lfloor n|w| \rfloor$ characters uniformly at random to replace. 
For each character, we fetch its $k$-th nearest neighbor from the model $\mathcal{M}$.
Finally, we apply these substitutions to the target word to obtain the perturbed word.

\subsection{Pairwise Comparisons}

We crowdsource legibility annotations for the perturbed words using Amazon's Mechanical Turk.
We collect annotations on both absolute legibility as well as relative 
preference between two differently perturbed inputs.
Since annotators tend to produce higher quality annotations when comparing items rather than assigning absolute values \cite{callison-burch-etal-2007-meta,liang2020beyond}, we design an annotation interface based on \textit{pairwise comparisons}
of two perturbed versions of the same word (\autoref{appendix:interface}).
Specifically, annotators see perturbations $w_1$, $w_2$ side-by-side, with the original word $w$ hidden.
They are asked to indicate which perturbation they find more legible by selecting
exactly one of these four labels:

\begin{description}
\addtolength{\itemindent}{1cm}
    \item[L1:] $w_1$ is preferred
    \item[L2:] $w_2$ is preferred
    \item[BL:] both $w_1$, $w_2$ are equally legible
    \item[NL:] neither $w_1$ nor $w_2$ is legible
\end{description}

$L_1$ and $L_2$ capture not only relative preferences between the two perturbations (used for the ranking task), but also indicate that the preferred perturbation is legible. However, these labels do not give us any information about the non-preferred perturbation.
On the other hand, the BL (Both Legible) and NL (Neither Legible) options do not give us a ranking between the two words,
but inform us about the legibility (or illegibility) of both words.
In the next section, we use these labels to derive datasets
for both a pairwise ranking task and a binary classification task.

We generate the data for annotation from English 
words consisting of the top $10,000$ frequent words (as per~\citet{10k_words}) 
in the Trillion Word Corpus \cite{brants2006web}. 
We filter this vocabulary to remove words with lengths less than $4$ or greater than $14$, ending up with $7600$ words. 
These words are randomly split into the train ($65\%$), validation ($15\%$), and test ($20\%$) sets; all future perturbation pairs $(w_1, w_2)$ generated for word $w$ are added to the corresponding set, and the same sets are used for all experiments.
To perturb a word $w$ into the pair $w_1$, $w_2$
a model $\mathcal{M}$ is picked at random from \{\trocr, \clip, \imgdot\} (the three best models from our initial
perturbation analysis).
We sample $k\sim \mathcal{N}(\mu_k, \sigma^2_k)$ and similarly for $n$, applying the appropriate bounds to keep $k>0$ and $n\in [0,1]$. The initial values are $\mu_k=25, \sigma^2_k=10, \mu_n=0.5, \sigma^2_n=0.2$.

\subsection{Adaptive Annotations}

The space of all possible perturbations of a word is vast,
and sampling the parameters $\phi$ based on the priors above
is unlikely to yield difficult perturbations which lie at the boundary
of legibility.
In order to identify such perturbations,
we collect data over multiple rounds using an \emph{adaptive}
process for generating the pairs.
In the first round, the pairs are generated as described above and
annotated by the crowd-workers.
In the following rounds, pairs are generated taking into account the last round of annotations.
Specifically, the $\phi_1, \phi_2$ for each successive round are chosen to make the next round of labeling harder for annotators. This is accomplished by manipulating the Gaussian used to generate $k, n$, i.e. by shifting $\mu_1, \mu_2$ to be closer to each other and reducing variance. 
This approach generates perturbations which elicit more nuanced comparisons from annotators, allowing us to capture fine-grained legibility preferences in the dataset.

\paragraph{Inter-annotator Agreement.} Three waves of annotations were collected using adaptive pair generation. To establish high quality and confidence in the test set labels, three annotations were collected for each pair of perturbations in the test set. Pairs where all annotators disagreed were removed from the test set.
For $49.1\%$ of pairs, all 3 annotators agree on the same label, $43.6\%$ of pairs have agreement between 2 out of the 3 annotators, and only $7.3\%$ of pairs have no agreement among annotators. Hence, even with $4$ labels to choose from, for $92.7\%$ of $(w_1, w_2)$ pairs, at least two out of three annotators chose the same label.
This suggests that the task is well-defined and has low variance.

\paragraph{Annotator Details.} We recruit $150$ annotators, all of whom had over a $95\%$ acceptance rate for previous work done on the platform, as well as a history of over $1,000$ completed tasks.\footnote{$686$ annotators were excluded due to failing their first quality check. Many attempts were observed to be spam.} Annotators are given occasional quality checks, wherein they annotate pairs drawn from a gold dataset labeled by the authors; annotators with less than $70\%$ accuracy on the gold data were removed from the study and their annotations discarded from the final dataset. Annotators are given batches of $20$ $(w_1, w_2)$ pairs at a time; typically taking between $30-45$ seconds to annotate. The average compensation per batch is $\$0.12$.
Further details of the annotation interface and instructions are 
available in \autoref{appendix:interface}.

\paragraph{Hard Subsets.} We identify challenging subsets of the collected data for the ranking and classification tasks. For ranking, the chosen subset ($N=1052$) contains pairs $(w_1, w_2)$ where $\dfrac{(n_1-n_2)^2}{n_1n_2} < 0.1$, 
i.e., both $n$’s are close to each other, so it is hard in the sense that the perturbations have similar parameters $\phi$ but varying degrees of legibility---they cannot be ranked just by comparing metadata. For the classification task, the chosen subset ($N=2626$) consists of all perturbations $w_i$ with $n_i>0.4$, making the task more challenging by excluding lightly-perturbed words which are easier to classify.

\begin{table}[!t]
\centering
\small
\begin{tabular}{@{}lrrrr@{}}
\toprule
 & \multicolumn{1}{r}{\# pairs} & \multicolumn{1}{r}{\# distinct} & \multicolumn{1}{r}{classification} & \multicolumn{1}{r}{ranking} \\
 & \multicolumn{1}{r}{$(w_1, w_2)$} & \multicolumn{1}{r}{$(w)$} & \multicolumn{1}{r}{examples} & \multicolumn{1}{r}{examples} \\ \midrule
Train & 14622 & 4940 & 20217 & 9027 \\
Val & 3326 & 1140 & 4639 & 2013 \\
Test & 3712 & 1520 & 4774 & 2650 \\ \midrule
Total & 21660 & 7600 & 29630 & 13690 \\ \bottomrule
\end{tabular}
\caption{ \dataset dataset statistics. For each word, there exist multiple perturbed pairs, generated through three rounds of adaptive annotations.}
\label{tab:dataset}
\end{table}

\section{Tasks and Models}
In this section, 
we start by introducing two tasks for
characterizing the legibility of perturbed texts,
followed by a number of models for solving them.

\subsection{Tasks}
\label{sec:tasks}

From the labels collected
in the previous section, we derive data for the two tasks: ranking and classification.
The tasks assume that the original word $w$ is known, as we base our setup considering an attacker who is trying to find the best perturbation.

\paragraph{Ranking Task.} Given a pair $(w_1, w_2)$ of perturbations and the original word $w$ as input, rank the perturbations in order of legibility. For this task, we only consider the subset of data labeled with strict rankings---i.e., excluding pairs labeled BL (Both Legible) and NL (Neither Legible).
As the data is balanced, we only report accuracy as the main metric for this task.

\paragraph{Classification Task.} Given a single perturbation $w_i$ and the original word $w$, decide whether the perturbation is legible. While annotators performed pairwise comparison between $(w_1, w_2)$, we can infer the binary legibility labels for $w_i$ from pairwise rankings as follows: for labels BL and NL, we can make the obvious inference of \emph{legible} and \emph{illegible} for both $w_i$. For labels $L_i$, we can again infer that $w_i$ is legible, but cannot say anything about $w_{j\neq i}$; all such $w_j$ with unknown legibility are excluded from the classification task dataset. Since there are more legible than illegible instances in the data,
we report both accuracy and F1 scores on this task.

\subsection{Baselines}

\paragraph{Majority Class.}
This baseline always predicts the majority class from the training set for every test example.
For the ranking task, it always predicts $w_2$ as the preferred perturbation
(resulting in an accuracy of $0.5$),
and for classification, the majority class is `legible' (yielding $0.677$ accuracy). 

\paragraph{Logistic Regression using $\phi$.}
Note that in an attack setting,
the attacker would know the perturbation
parameters $\phi$ exactly and may be interested in predicting the legibility
of their perturbation using these parameters.
Hence,
we perform logistic regression directly on the attack parameters ($n,k$) to predict the label. 
Being a simple metadata-only baseline, this model does not take into account the characters that were perturbed
or their position. 
\begin{figure*}[t]
    \centering
    \includegraphics[width=5.5in]{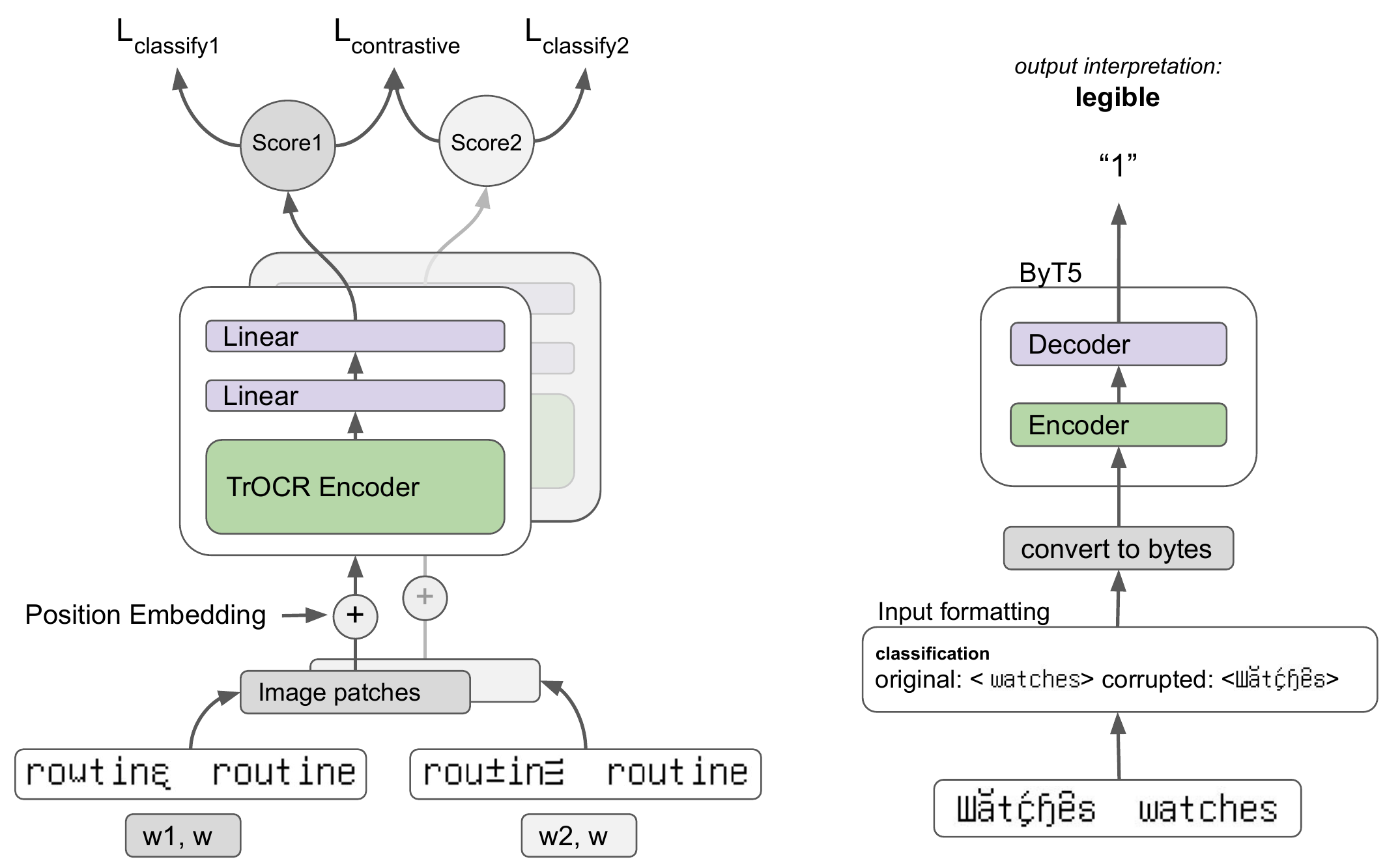}
    \caption{Comparing ByT5 and TrOCR training setup. ByT5: Both the perturbed and original words are given as one input to the model. \trocr: Both $w_1, w_2$ are fed sequentially into the same TrOCR-based model, and the two resulting scalar outputs are used to compute the loss. For each perturbation, the string ``$w_i$ $w$" is rendered and used as input for the model.}
    \label{fig:models}
\end{figure*}
\subsection{Text-based Models}
\paragraph{ByT5.} Legibility, as defined in this paper,
is a \textit{visual} property.
However, we might expect pretrained language representations
(e.g., those learned by large-scale language models)
to also encode visual similarity between characters
since the web-corpora used for pretraining might include similar-looking
characters in the same contexts (e.g., `0' instead of `O').
To test this, we experiment
with ByT5 \cite{xue_byt5_2021},
a multilingual encoder-decoder language model which tokenizes
inputs into byte sequences.
Byte-level tokenization ensures that none of the perturbations
in \dataset are out-of-vocabulary,
and multilingual pretraining ensures that the model
has seen a large subset of Unicode.
We finetune the pretrained ByT5-models (`small' and `base') to predict 
the binary labels for both classification and ranking in a text-in text-out setting.
For ranking, the inputs are formatted
as: 
``original: $<w>$ word0: $<w_0>$ word1: $<w_1>$'',
and the output is ``0'' or ``1'' depending on which word is more legible.
For classification,
the inputs are formatted as:
``original: $<w>$ corrupted: $<w_i>$ '',
and the output is ``0'' or ``1'' depending on whether the corruption
is illegible or legible.
We train two separate models starting from the pretrained ByT5 weights
using the cross-entropy loss over the target byte-sequence
and AdamW optimizer \cite{adamw}
and perform early stopping using the validation set. \autoref{fig:models} outlines the model schematic with sample inputs and outputs.
\subsection{Vision-based Models}
Since we are concerned with finding representation spaces for \emph{visually} similar characters, vision-based models are a natural choice for the task.
We consider both unsupervised models which rely on pixel-based or
embedding-based similarities,
as well as supervised models based on OCR, which we train on the \dataset data.

\paragraph{\imgdot.}
This unsupervised approach compares the corresponding characters in $w$ and $w_i$
based on the cosine distance between their pixel renderings.
For the ranking task, this model selects the perturbation whose average cosine
distance with the uncorrupted word is lower.
For classification,
we tune a threshold similarity parameter on the training set,  above which the model predicts
`legible'.

\paragraph{\trocr-Embeddings.} This approach is identical to \imgdot,
except that we use the pretrained character embeddings obtained by passing the
rendered images as input to the \trocr model. The embedding vector for each character is obtained by averaging the last hidden state from the encoder output (bypassing the pooler).
Note that the accuracy of these unsupervised baselines gives us an idea of how well the corresponding representations align with human notions of legibility.

\paragraph{\trocr.}
Finally, we consider finetuning \trocr on the \dataset data.
We only use the encoder part of the \trocr base model
and connect it to a linear head. This linear head has two fully connected layers mapping inputs of size $768$ (which is equal to the dimension of the encoder output) to a scalar output
which represents the legibility score of the perturbed input.
We use ReLU activations between the linear layers
and apply dropout.
The model takes variable-sized images as input; this is created by rendering a pair $(w_i, w)$ into a single image by concatenating both strings along the horizontal axis (see \autoref{fig:models}). 

For the classification task, the output score from the model
is used directly for predicting the label.
Given a pair $(w, w_i)$, let $s_i$ denote the scalar output
from the model and let $y_i\in \{0, 1\}$ denote the legibility label
(where $1$ denotes that $w_i$ is legible).
Then the classification loss is given by:
\begin{align}
    \mathcal{L}_{\text{classify-}i} = - & y_i \log \sigma(s_i) \nonumber \\ & - (1-y_i) \log \left[1 - \sigma(s_i)\right]
\end{align}
where $\sigma$ is the sigmoid function.
We apply the same loss function to both perturbations
$w_1$ and $w_2$.
We denote this classification model as \trocr-C.

For the ranking task,
we use the same model but apply it separately to the pairs
$(w, w_1)$ and $(w, w_2)$ to obtain the scores $s_1$ and $s_2$.
The parameters across the two applications of the model are shared
in a Siamese network setup \cite{koch2015siamese}.
Given these two scores, and the label $y \in \{0, 1\}$
(where $0$ denotes that $w_1$ is more legible),
we define the ranking loss as:
\begin{align}
    \mathcal{L}_\text{contrastive} = -&y \log \sigma(s_1 - s_2) \nonumber \\ 
    & - (1- y) \log \left[1 - \sigma(s_1 - s_2)\right]
\end{align}
The above loss encourages $s_1$ to be higher than $s_2$ when $y=0$
and vice versa.
A similar loss has been used to train summarization models
from pairwise human preferences \cite{stiennon2020learning}.
We denote this ranking model as \trocr-R.

The Siamese setup for the ranking task is limited in the sense
that it cannot directly compare the two perturbations to decide
which is more legible.
However, our goal is to train the model to produce
a calibrated legibility score given only a single perturbation as
the input.
Further, the Siamese network allows us to train the model
on \textit{both} the classification and ranking tasks
together in a multi-task fashion:
\begin{equation}
    \mathcal{L} = \mathcal{L}_{\text{classify-}1} + \mathcal{L}_{\text{classify-}2} + \mathcal{L}_{\text{contrastive}}
\end{equation}

The loss terms for each training example are masked based on the label: the ranking loss is masked out if the label is ``equally legible" or ``both unclear", whereas the individual classify-$i$ loss is masked out if the inferred binary legibility of perturbation $w_i$ is indeterminate (e.g. for label $L_1$, binary legibility of $w_2$ is unknown).
Together, these losses ensure that the legibility score $s_i$
is thresholded at $0$, above which the perturbations are legible,
and more legible inputs receive a higher score.
We denote the model using combined loss as \trocr-MT.

\section{Results}
\label{sec:results}
\begin{figure}[t]
    \centering
    \includegraphics[height=2.7in]{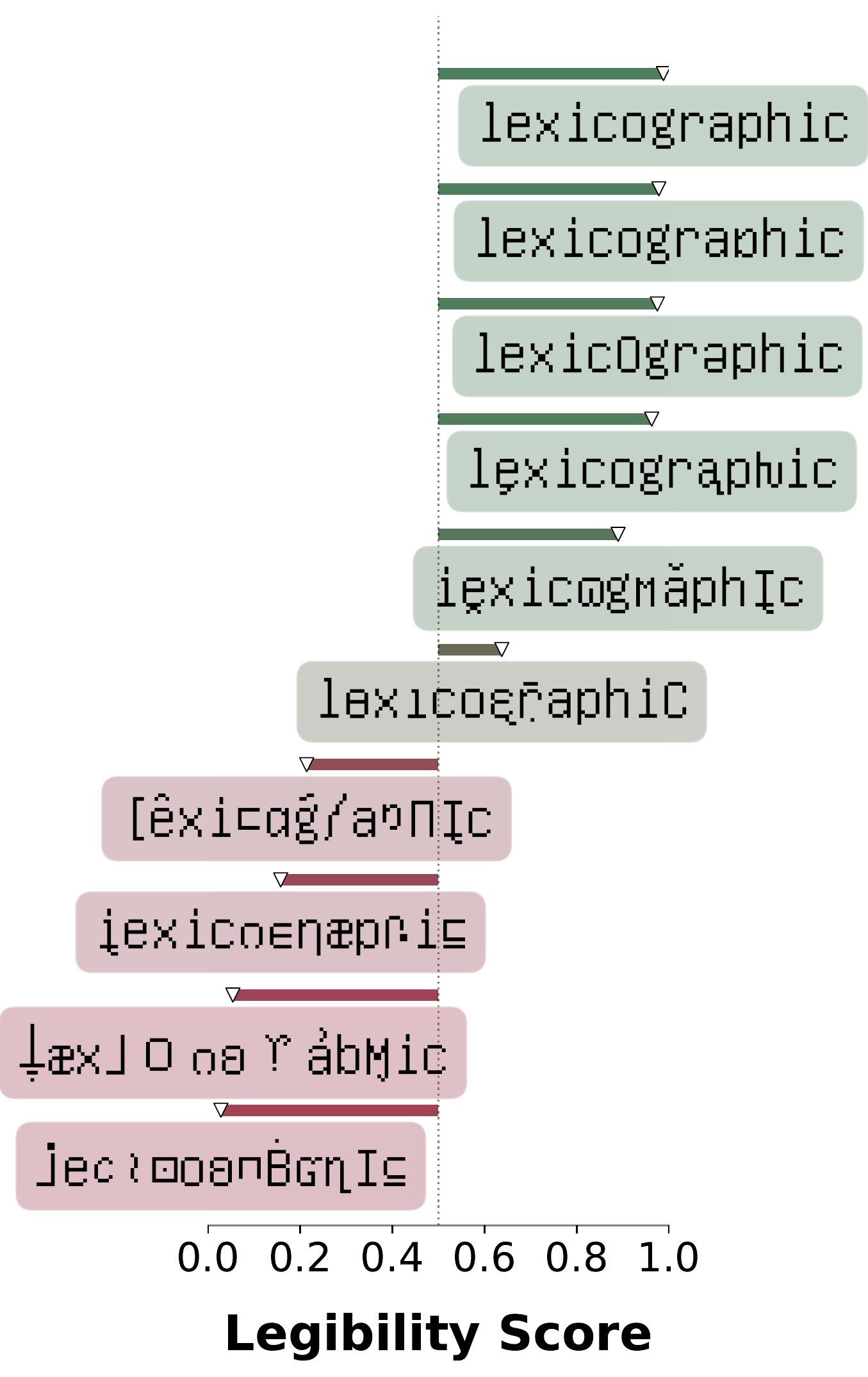}
    \includegraphics[height=2.7in]{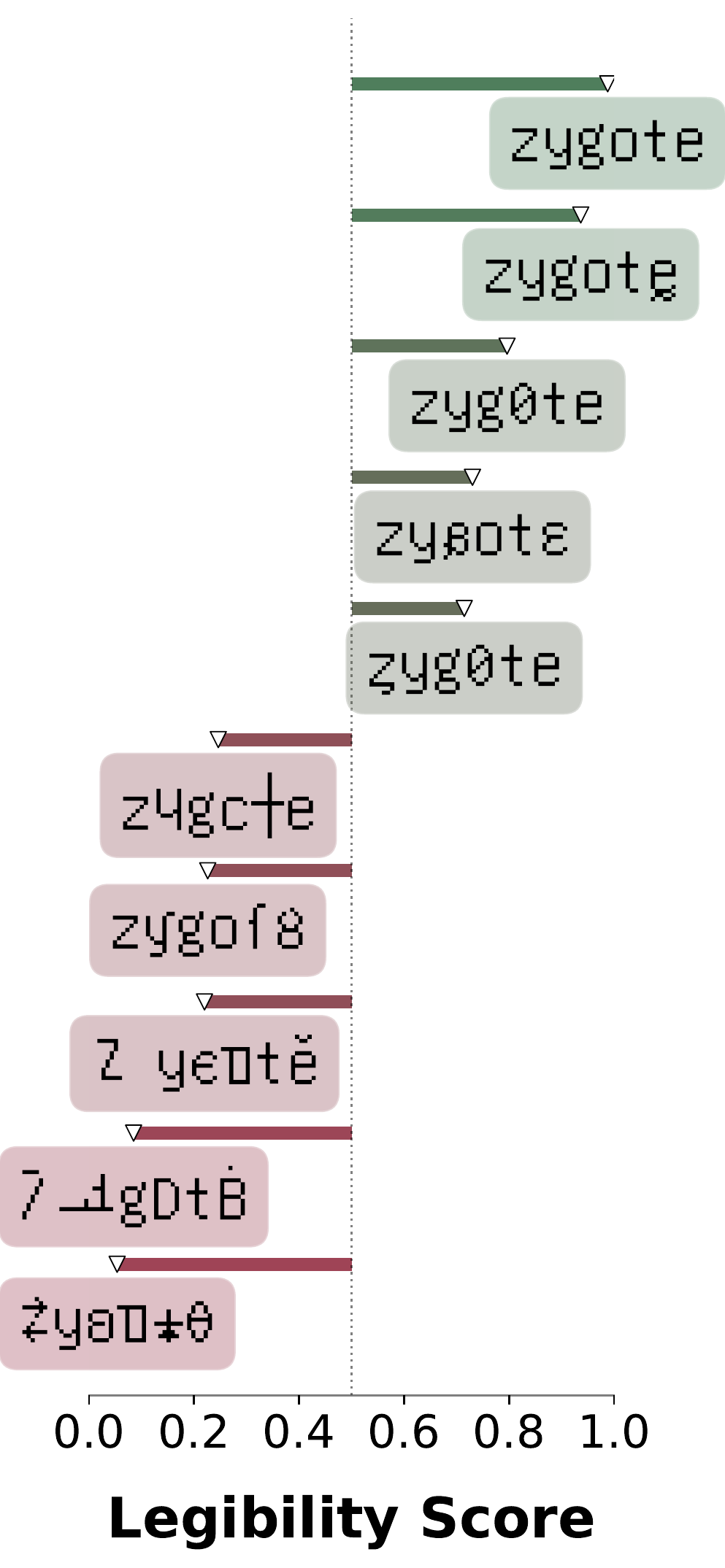}
    \caption{Legibility scores for \dataset-generated perturbations of \emph{lexicographic} and \emph{zygote} from the \trocr-MT model. Neither word was seen during training.
    }
    \label{fig:scores}
\end{figure}

\begin{table*}[ht]
\centering
\small{
\begin{tabular}{@{}lllrrrrr@{}}
\toprule
\multicolumn{2}{l}{\multirow{1}{*}{}}
& \multicolumn{1}{l}{\multirow{1}{*}{\textbf{Model}}} & \multicolumn{4}{c}{Classification} & \multicolumn{1}{c}{Ranking} \\ \cmidrule(lr){4-7} \cmidrule(lr){8-8}
\multicolumn{2}{c}{} & \multicolumn{1}{c}{} & \multicolumn{1}{l}{\textbf{Accuracy}} & \multicolumn{1}{l}{\textbf{Precision}} & \multicolumn{1}{l}{\textbf{Recall}} & \multicolumn{1}{l}{\textbf{F1} Std/Hard} & \multicolumn{1}{r}{\textbf{Accuracy} Std/Hard} \\ \midrule
\multirow{4}{*}{} Baselines & \multirow{1}{*}{} & Majority Class & $.512$ & $.512$ & $1.000$ & $.677 / .000$ & $.500/.502$ \\
 &  & Log. Regression & $.680$ & .$659$ & $.671$ & $.665/.256$ & $.744/.642$ \\
\midrule
 Vision-based & \multirow{4}{*}{}  & \imgdot & $.788$ & $.861$ & $.828$ & $.845/.583$ & $.790/.652$ \\
 &  & TrOCR embeds & $.825$ & $.868$ & $.883$ & $.868/.654$ & $.781/.677$ \\
 &  & TrOCR-C & $.840$ & $.881$ & $.891$ & $.886/$\phantom{$.1$}--\phantom{$5$} & \multicolumn{1}{r}{\phantom{$.581$}--\phantom{$14$}} \\
 &  & TrOCR-R & \multicolumn{1}{r}{\phantom{.5}--\phantom{1}} & \multicolumn{1}{r}{\phantom{.5}--\phantom{1}} & \multicolumn{1}{r}{\phantom{.5}--\phantom{1}} & \multicolumn{1}{r}{\phantom{.581}--\phantom{.134}} & $.835/$\phantom{$.1$}--\phantom{$5$} \\
 \multirow{4}{*}{} &  & TrOCR-MT & $\mathbf{.868}$ & $\mathbf{.914}$ & $.895$ & $\mathbf{.905/.726}$ & $\mathbf{.858/.757}$ \\
   \midrule 
  Text-based & \multirow{1}{*}{}  & ByT5-small & $.844$ & $.872$ & $.909$ & $.890/$\phantom{$.1$}--\phantom{$5$} & $.762/$\phantom{$.1$}--\phantom{$5$} \\
 &  & ByT5-base & $.842$ & $.868$ & $\mathbf{.912}$ & $.889/$\phantom{$.1$}--\phantom{$5$} & $.769/$\phantom{$.1$}--\phantom{$5$} \\
 \bottomrule
\end{tabular}
}
\caption{Results on the standard test set. The TrOCR-MT model, trained in the multi-task setting, outperforms all other models for F1 score on both tasks. The trained models also outperform the baselines on both tasks.
}
    \label{tab:results}
\end{table*}
\autoref{tab:results} shows the performance
of all models introduced on both the classification
and ranking tasks.

\begin{figure}
    \centering
    \includegraphics[width=1.49in]{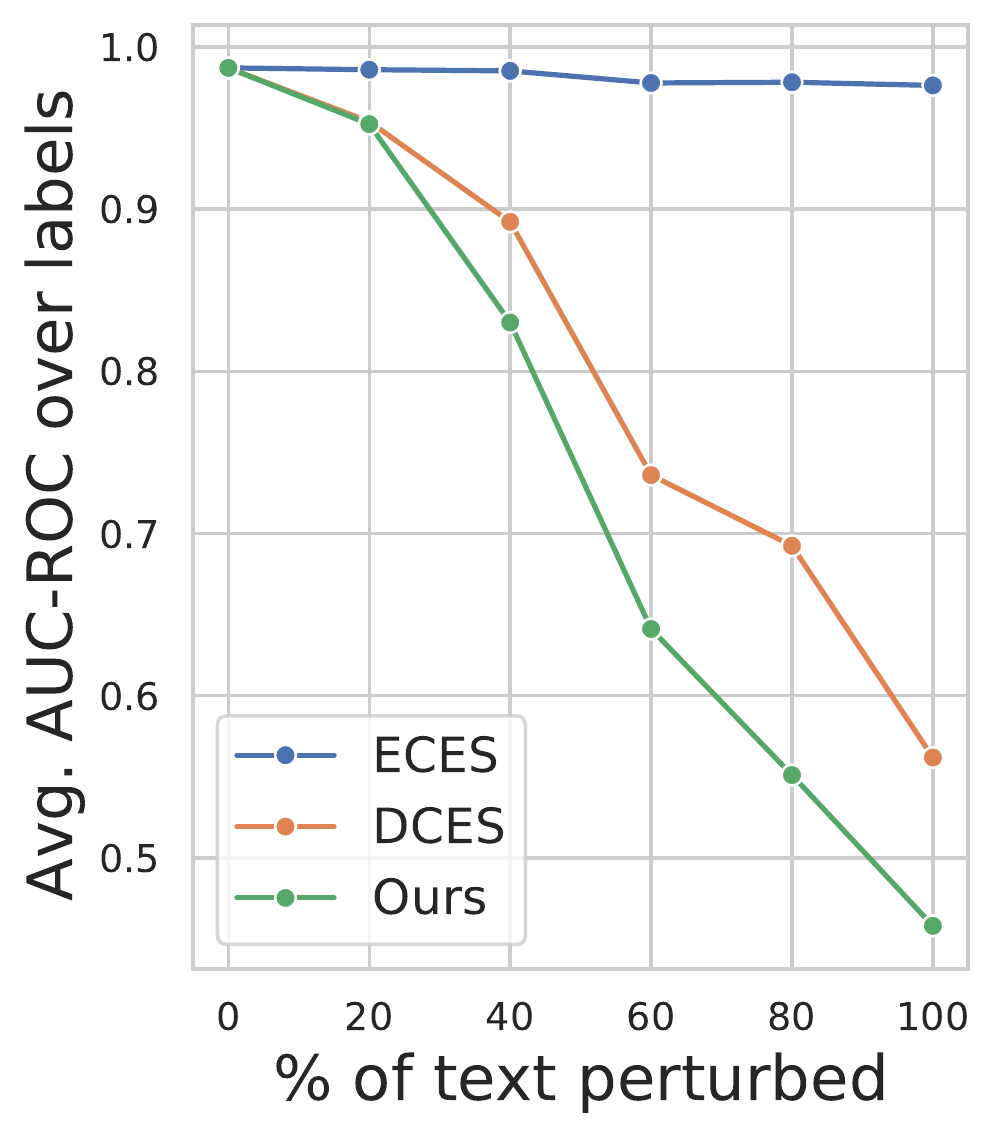}
    \includegraphics[width=1.5in]{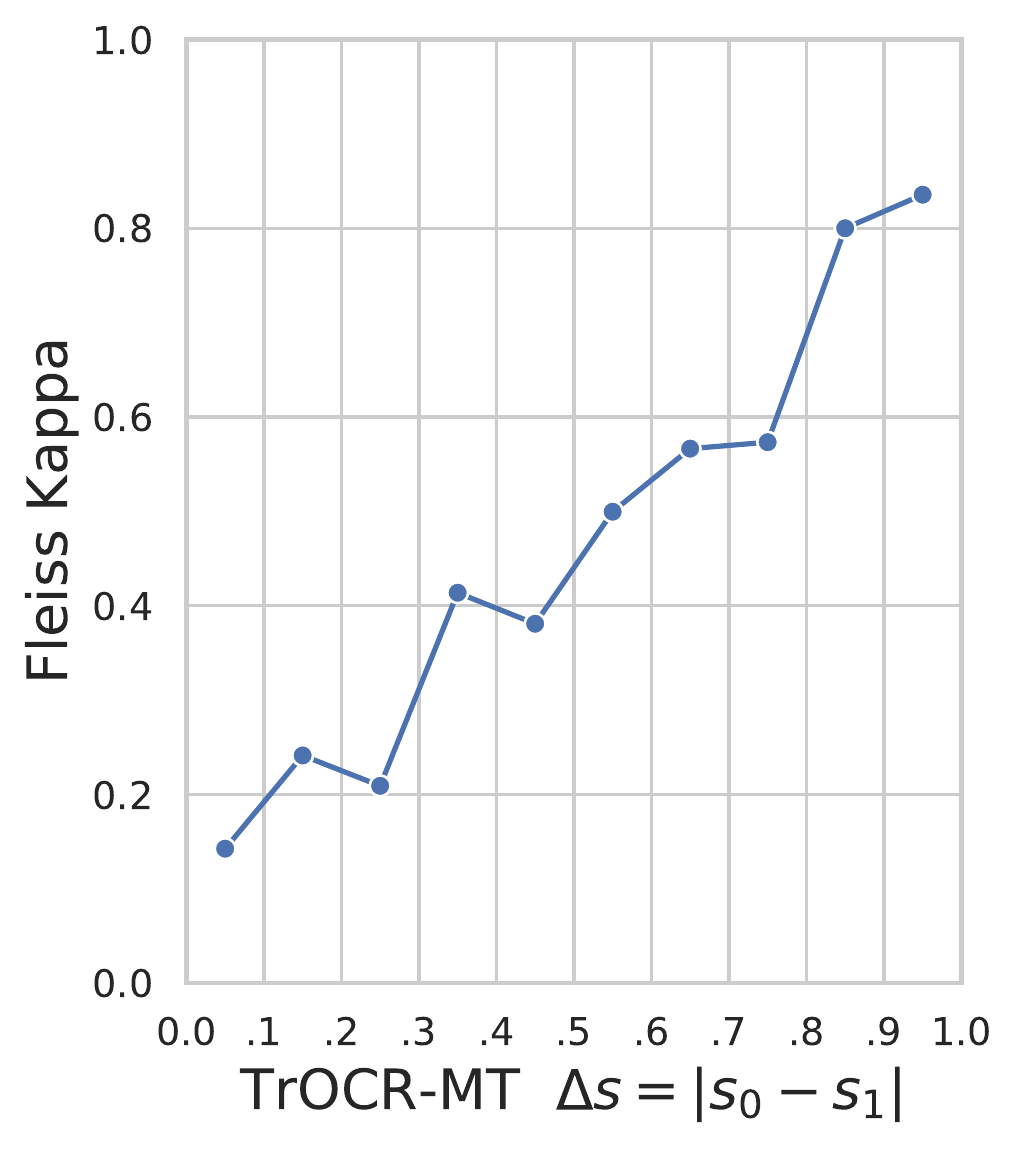}
    \caption{
    \textbf{Left}:
    \emph{Detoxify} model performance on perturbations
    generated by different attack methods on a random subset ($N=2000$) of the Jigsaw Toxic Comment Classification dataset. Model performance degrades most on our perturbations. \textbf{Right}: Model confidence on legibility is aligned with annotator agreement. Legibility scores $(s_0,s_1)$ were obtained using \trocr-MT for each perturbed pair $(w_0,w_1)$ in the test set. Pairs were grouped by the score difference $\Delta s=|s_0-s_1|$ and Fleiss’ $\kappa$ was computed 
    for each group.
    }
    \label{fig:jigsaw}
\end{figure}

\paragraph{Classification Task.}
For the classification task, 
we find that baselines that just use the  metadata perform poorly. The Majority Class baseline obtains an F1 score of $0.677$, and the Logistic Regression model using $\phi$ parameters yields an F1 score of $0.665$, implying that legibility is \emph{not} a simple function of the perturbation parameters $k,n$.
The unsupervised vision-based models, \imgdot and \trocr embeddings, vastly improve upon the simple baselines, with the \trocr embeddings obtaining an F1 score $0.868$ and \imgdot yielding an F1 score of $0.845$. 
Hence, these embeddings align reasonably well with human
perceptions of legibility.
The text-based ByT5 models improve significantly over the baselines and
unsupervised vision-based models.
They are
comparable to the performance of the single-task objective \trocr-C, but worse than the \trocr-MT. 
This suggests that the ByT5 models might have encountered
some visual perturbations
during pretraining.
Comparing the single-task \trocr-C model with the multi-task \trocr-MT, we find that the presence of the additional ranking loss term during training improves model performance on the classification task from $0.886$ to $0.905$. 
On test examples where all $3$ annotators agree, \trocr performs even better, attaining an F1 score of $0.960$, compared to a score of $0.850$ on examples where only $2$ annotators agree. 
As further evidence of the model's alignment with annotators, we find that the model confidence is directly correlated with annotator agreement (cf. \autoref{fig:jigsaw}) as measured by Fleiss' $\kappa$ \cite{fleiss1971measuring}. Furthermore, consider \autoref{fig:scores}, which shows legibility scores obtained from \trocr-MT for two words picked at random which are not part of the training set. Qualitatively, we see that legibility scores from the \trocr-MT model 
aligns with the human judgements of legibility for these words. 

\paragraph{Ranking Task.}
The \trocr-MT model performs better \emph{relative to} other models, resulting in a $6.8\%$ absolute accuracy improvement.
Akin to the classification task, we find the \trocr-MT model outperforms its single-task counterpart \trocr-R. Thus, training with a multi-task objective improves performance on both ranking and classification tasks when compared to single-objective models.
Differently from classification,
we find that ByT5 is significantly worse than the vision-based models
on ranking,
suggesting that language model pretraining is effective at
separating legible from illegible perturbations,
but not at encoding the degree of legibility
of legible perturbations.

\paragraph{Jigsaw Challenge.}
Next, we check whether perturbations generated
by our attack model (\S~\ref{sec:perturbation})
and filtered to ensure legibility using \trocr-MT
are effective at degrading the performance of NLP models.
We employ the Jigsaw Toxic Comment Classification Dataset, which is a multilabel classification dataset consisting of Wikipedia comments and human-annotated binary labels for 6 toxicity categories.
In \autoref{fig:jigsaw} (Left), we compare \dataset and VIPER-DCES strategies in a real-world scenario by perturbing the Jigsaw dataset with each strategy and reporting how much these perturbations degrade the performance of \texttt{Detoxify-original} \cite{Detoxify}, a BERT-based model which has state-of-the-art performance on the Jigsaw dataset. We show that \dataset produces greater degradation at lower $n$, and produces more legible perturbations even at higher $n$ (due to \trocr-MT filtering).
In comparison, we find that DCES perturbations become very hard to read at higher $n$, diluting the significance of the DCES results at high $n$. \autoref{appendix:toxic} provides a qualitative analysis of the legibility of DCES perturbations compared to those generated using our \dataset method.

VIPER-ECES 
causes a negligible degradation on model performance, which is due to the fact that the BERT tokenizer ``corrects" almost all of the simple diacritic-based ECES character substitutions. This means that the classification model receives mostly unperturbed input save for some isolated UNKs. For example, the perturbed input \texttt{Ťĥâňǩ ŷôǔ} is tokenized back into \texttt{Thank you}.
Taken together, these results demonstrate that \dataset exploits a more \emph{efficient} legibility space, finding character substitutions which have a greater impact on model performance while preserving legibility.

\begin{figure}
    \centering
\includegraphics[height=1.8in]{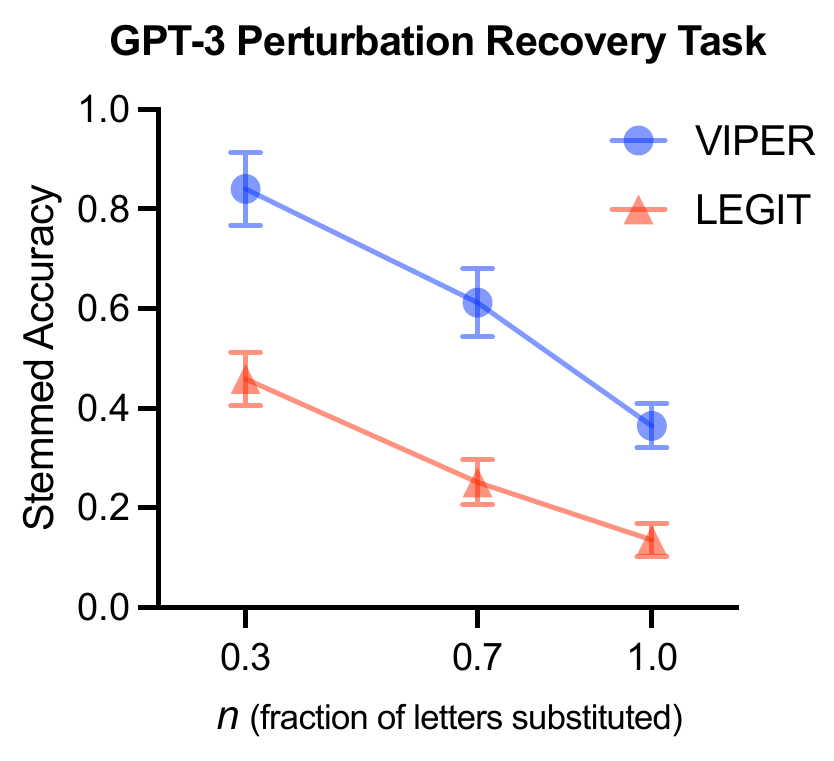}
    \caption{ \dataset perturbations sampled at low $n$ degrade accuracy at levels comparable to the $n=1$ VIPER configuration.
    Error bars indicate 95\% confidence interval.
    }
    \label{fig:gpt3}
\end{figure}

\paragraph{Perturbing GPT-3.}
The strong performance of ByT5 at separating legible
from illegible inputs suggests that language models
might be somewhat robust to such perturbations.
To examine this, we experiment with GPT-3 (\texttt{text-davinci-002} checkpoint) \cite{GPT3}
using a 
\emph{perturbation recovery task}, wherein we prompt the model to decode perturbed words back to their original strings.
We sample a subset of $1,000$ ($w, w_i$) pairs from \dataset which have a label of \emph{legible}.
These perturbations are fed to the GPT-$3$ model in batches of $10$, along with an instructional prompt (see \autoref{appendix:gpt3}) and $4$ examples; recovered words are received as a completion to the input prompt.
In addition, we also perturb the same $1000$ words using VIPER-DCES and report the accuracy of GPT-3 at reconstructing them.
We observe that GPT-$3$ often returns a word with a short edit distance to the original word,
and hence to 
capture this in our evaluation, we apply the Porter stemmer from NLTK \cite{journals/corr/cs-CL-0205028} to both the original words and predicted reconstructions, and then measure how often their stemmed forms are the same. We repeat this experiment $3$ times, randomly sampling the $4$ examples in the prompt each time.
\autoref{fig:gpt3} shows the GPT-3 accuracy at different fractions of corrupted characters
($n=\{0.3, 0.7, 1.0\}$).
As expected, the accuracy goes down as $n$ increases,
but we find that GPT-3 performs worse on \dataset perturbations.
This demonstrates that while state-of-the-art language models are mildly robust to the narrower range of perturbations considered in existing visual attacks, they degrade significantly on inputs sampled from \dataset which are marked by humans as legible.
This result underscores the importance of considering the entire space of legible perturbations when evaluating model robustness.

\section{Conclusion}
We set out to 
characterize the limits of legibility of visual perturbations. 
To do so, 
we first collected and released
a new dataset, \dataset, comprising
legibility preferences of human subjects.
Using this dataset, 
we framed 
a binary legible-or-not classification task, and a ranking task to rank candidate perturbations.
For these tasks, we explored several text- and vision-based models, and found that 
our models 
obtain a high F1 score of 0.91 for the classification task and an accuracy of 
0.86 for the ranking task.
Perturbations generated using the same attack method as used
for constructing \dataset lead to significant degradation
on the Jigsaw Challenge task and are not recovered by GPT-3 accurately,
despite being filtered for legibility.
We believe this work 
opens avenues for research on
legibility-driven certified robustness
to visual attacks in NLP.

\section*{Limitations}
At the outset, we note that while our legibility-scoring models are a step forward towards defending against visual attacks, they should not be seen as perfect. Defending against all of the attacks which our models find legible might still leave room for legible attacks missed by our system.

Moreover, we note that the perturbation procedure outlined here only generates
substitution-based perturbations.
Whereas, characters may also be
deleted, added, or swapped, and multiple adjacent characters may be
substituted with visually similar counterparts
(see \autoref{fig:manipulated_examples}).
Future work may explore broader classes of perturbations.

When constructing the dataset, we only chose words with a length of at least 4 letters, excluding many common 3-letter words. This is because for 3-letter words, there is a high likelihood that a bad perturbation may be mistakenly recognized as a good perturbation by virtue of being in-vocabulary. For example, ``ban'' is a bad perturbation of ``man'', but for an annotator who sees it without knowing the original word and in absence of any sentence-level context, it seems like a perfectly good perturbation, when in fact it obscures the meaning of the original word. This is a limitation of the experimental setup that can lead to bad annotations, and to mitigate it we chose a higher minimum word length as longer words have fewer such collisions. 

Further, we study word-level perturbations in isolation 
without any surrounding context, whereas 
in practice, readers often can 
decipher words based on the context. 
In general, the legibility of a text depends on the context
around it---for example,
even if a word is deleted from a sentence it is often possible to  reconstruct it. 
The data we collect here, however,
measures the legibility of individual words without any context,
in order to simplify the generation and annotation process.
As a result, the legibility estimated using this data should
be considered as \textit{lower bound} of the legibility in any given context. 
This was a deliberate choice as 
we wanted to ensure that 
whatever we ascertain as legible 
is legible in \emph{all} contexts.

Lastly, the models we develop in our work 
are of relatively moderate size ($334 - 584$ million parameters) and take only unimodal input (i.e. pixels for \trocr models and Unicode bytes for ByT5).
and future work may be able
to improve the performance by using larger models which accept multimodal input (e.g. both pixels and Unicode bytes simultaneously) and learn joint representations across these modalities.

\section*{Ethical Considerations}
The word list comprising our dataset was filtered to remove swear words, slurs etc. in order to avoid exposing annotators to potentially harmful content.

\section*{Acknowledgements}
We thank Professors Carlo Tomasi and Sam Wiseman at Duke University for their helpful feedback. This research was supported by a grant from the Arts \& Sciences Council Committee on Faculty Research at Duke University.

\bibliography{anthology,custom}
\bibliographystyle{acl_natbib}

\appendix
\section{mTurk Annotation Inferface}
\label{appendix:interface}
\begin{figure*}[ht]
    \centering
    \includegraphics[width=0.9\textwidth]{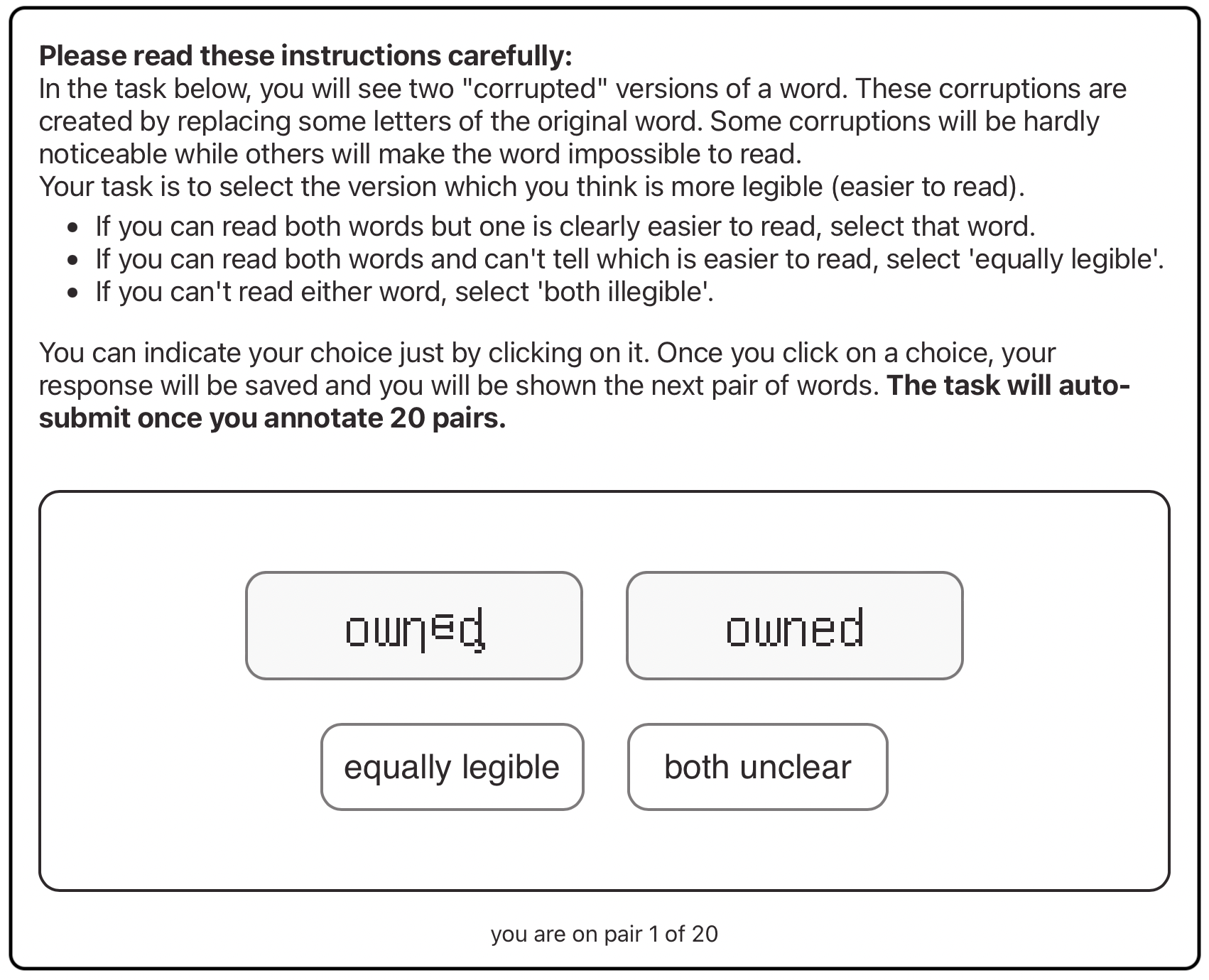}
    \caption{The mTurk Annotation Interface}
    \label{fig:my_label}
\end{figure*}

The web-based UI used by mTurk Annotators is shown in \autoref{fig:my_label}, with the instructions being visible throughout the duration of the task. Note that the perturbed words $w_1, w_2$ are rendered in GNU Unifont, which is the same font that words are rendered in for computing visual similarity (cf. \S Legibility Tests, Perturbation Process). This ensures that both annotators and visual similarity models see pixel-for-pixel identical perturbations, controlling for the fact that different fonts render the same character differently.

The interface is optimized for clarity and labeling speed, with a focus on eliminating unnecessary UI elements and minimizing clicks. Labeling each pair $w_1, w_2$ with one label $L\in\{L_1, L_2, BL, NL\}$ takes exactly one click. Annotators choose $L_1$ by clicking on $w_1$ (the left word), and similarly by clicking on $w_2$ (the right word) for $w_2$. $BL$ is selected using the ``equally legible" button, whereas $NL$ is chosen by clicking on ``both unclear."

Immediately after a choice is made, the UI updates and the next pair in the batch is shown (there is no option to go back and edit the chosen label). Annotators who attempt to cheat on the task by ``speeding through" (i.e. clicking randomly or spamming the same choice) end up failing the occasionally administered quality checks and are subsequently disinvited from the study.

\section{Use of OCR Models}
\label{appendix:OCR}
\citet{boucher_bad_2021} propose using Optical Character Recognition (OCR) models to preprocess input for text-based language models. Rendering input text and passing it through an OCR before giving it to the language model filters certain kinds of misleading Unicode characters (e.g. invisible control sequences or near-identical Confusables \cite{davis_uts_2021}) from the text. However, when used for legible but visually distinct perturbations, off-the-shelf OCR models run into two problems. 

Firstly, both mono- and multi-lingual OCR models will recognize characters from learned scripts at face value, instead of recognizing their intended use as visually similar substitutions. For example, \trocr \cite{li_trocr_2021}, when given an image of the string `Mex!(0', decodes it into `Mex!(0' (i.e. the same string), completely ignoring its intended meaning (Mexico). Secondly, since OCR models are only trained on semantically meaningful inputs, they do not learn good priors to differentiate non-sense inputs from highly perturbed inputs.

We use two OCR-capable models on the ranking and classification tasks: \trocr, which is explicitly trained on an OCR dataset, as well as CLIP, which is trained on a general corpus containing images of texts
from which it learns ``a high quality semantic OCR representation that performs well on digitally rendered text''
\cite{radford_learning_2021}.

We find that \trocr models fine-tuned on our dataset achieve high performance on legibility-related tasks.
On the text side, we consider the token-free language model, ByT5 \cite{xue_byt5_2021}, which encodes each byte individually, as opposed to byte-pairs or subword tokens longer than one byte. Since its encoding of each byte is disentangled from surrounding bytes, ByT5 is able to retain a larger share of the unperturbed part of the string, hopefully making it more robust to character-substitution perturbations compared to token-based models, which reduce the sequences with perturbed characters into rare tokens or simply to \texttt{UNK}s. 

\section{Hyperparameters}
\label{appendix:hyperparameters}
\trocr (`base-handwritten' version) was fine-tuned on \dataset with the loss function configurations (C, R, MT) described above. To train each configuration, we use a single NVIDIA A$6000$ GPU ($48$GB VRAM) with a batch size of $26$ and learning rate of $10^{-5}$ with the AdamW optimizer and a linear decay schedule (without warmup).
ByT5-base and ByT5-small were trained on the same hardware with a batch size of $8$ and learning rate of $10^{-4}$.

\section{Toxic Comment Classification Experiment}
\label{appendix:toxic}
The original string from the Jigsaw Toxic Comment Classification dataset is:

\begin{quote}
\texttt{
It is needed in this case to clarify that UB is a SUNY Center. It says it even in Binghampton University at Albany, State University of New York, and Stony Brook University. Stop trying to say it's not because I am totally right in this case.
}
\end{quote}

\begin{figure*}[h]
    \centering
    \begin{subfigure}[h]{\textwidth}
    \centering
        \includegraphics[width=5in]{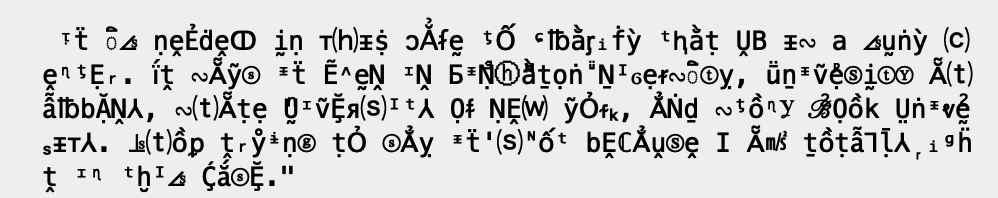}
        \caption{VIPER DCES, $n=1.0$, nearest neighbors sampled uniformly from list of top 10 neighbors for each character.}
    \end{subfigure}
    \begin{subfigure}[h]{\textwidth}
    \centering
        \includegraphics[width=5in]{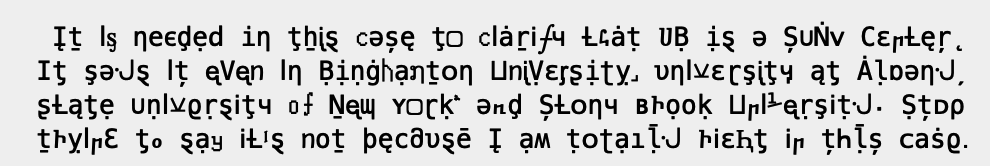}
        \caption{Ours (\dataset perturbation strategy with \trocr-MT legibility filter) $n=1.0$, nearest neighbors sampled normally ($\mu=15$, $\sigma^2=7$) from top 30 neighbors for each character}
    \end{subfigure}
    \caption{A randomly selected paragraph from the Jigsaw dataset (a) perturbed by VIPER DCES (b) and our method (c). Our perturbation appears more legible despite being generated using harsher parameters.} 
    \label{fig:randomjigsaw}
\end{figure*}

The VIPER DCES and \dataset perturbations are compared in \autoref{fig:randomjigsaw}. The \dataset perturbations were labeled as legible by \trocr-MT.

\section{GPT-3 Experiment}
\label{appendix:gpt3}
We provided the following prompt to the \texttt{text-davinci-002} checkpoint using the GPT-3 API:
\begin{quote}
\texttt{
The following is a list of corrupted words and their correct versions. The corruptions were created by replacing some or all letters of the correct version with similar-looking letters. \\
Corrupted: \\
1. $c_1$ \\
2. $c_2$ \\ 
... \\
10. $c_{10}$ \\
Original: \\
1. $o_1$ \\
2. $o_2$ \\ 
3. $o_3$ \\
4. $o_4$ \\
5.
}
\end{quote}
The model is allowed to condition on 4 ground-truth examples: $o_1$ through $o_4$, and attempts to generate $o_5$ through $o_{10}$ by providing a completion for the prompt above. The \emph{temperature} and \emph{top p} parameters were both set to 1 to allow for consistent and reproducible outputs across batches.

\end{document}